\documentclass[runningheads]{llncs}
\usepackage{graphicx}
\usepackage{hyperref}
\usepackage{amsmath}
\usepackage{array}
\usepackage{amsfonts}
\usepackage{booktabs}
\usepackage{stmaryrd}
\usepackage{siunitx}

\usepackage{float}
\usepackage{multirow}
\usepackage{ragged2e}

\usepackage{enumitem}
\usepackage[utf8]{inputenc}
\usepackage{xcolor}
\usepackage{courier}
\newcommand{\ct}[1]{\texttt{#1}}
\usepackage{algorithm,algpseudocode}
\newcolumntype{P}[1]{>{\centering\arraybackslash}p{#1}}
\newcolumntype{M}[1]{>{\centering\arraybackslash}m{#1}}

\DeclareMathOperator{\mP}{\mathcal{P}}
\DeclareMathOperator{\tP}{\tilde{\mathcal{P}}}
\begin{document}

\title{\ct{FedSLD}: Federated Learning with Shared Label Distribution for Medical Image Classification}
\author{Jun Luo\inst{1} \and
Shandong Wu\inst{1,2}}
\authorrunning{J. Luo and S. Wu}
\institute{Intelligent Systems Program, University of Pittsburgh, Pittsburgh, PA, USA, 15213\\
\email{jul117@pitt.edu} \and
Department of Radiology, Department of Biomedical Informatics, and Department of Bioengineering, University of Pittsburgh, Pittsburgh, PA, USA, 15213\\\email{wus3@upmc.edu}}
\maketitle              % typeset the header of the contribution

\begin{abstract}
Machine learning in medical research, by nature, needs careful attention on obeying the regulations of data privacy, making it difficult to train a machine learning model over gathered data from different medical centers. Failure of leveraging data of the same kind may result in poor generalizability for the trained model. Federated learning (FL) enables collaboratively training a joint model while keeping the data decentralized for multiple medical centers. However, federated optimizations often suffer from the heterogeneity of the data distribution across medical centers. In this work, we propose Federated Learning with Shared Label Distribution (\ct{FedSLD}) for classification tasks, a method that assumes knowledge of the label distributions for all the participating clients in the federation. \ct{FedSLD} adjusts the contribution of each data sample to the local objective during optimization given knowledge of the distribution, mitigating the instability brought by data heterogeneity across all clients. We conduct extensive experiments on four publicly available image datasets with different types of non-IID data distributions. Our results show that \ct{FedSLD} achieves better convergence performance than the compared leading FL optimization algorithms, increasing the test accuracy by up to 5.50 percentage points.
\end{abstract}
\section{Introduction}
\label{Introduction}
Deep learning (DL) is well known for requiring a large amount of data for robust training of generalizable models. For DL in medical research~\cite{wang2016deep,oh2020deep,lee2017deep,rajpurkar2017chexnet}, large datasets can be difficult to obtain since the data collected by medical centers and hospitals are often privacy-sensitive. Therefore, sharing of the raw data between institutions is usually constrained by the restrictions such as Health Insurance Portability and Accountability Act (HIPAA) in the United States, and General Data Protection Regulation (GDPR) in Europe.

The recent emergence of federated learning (FL)~\cite{mcmahan2017communication,kairouz2019advances,li2020federated} has provided this issue with a feasible solution. FL is a distributed machine learning scenario where only the model weights are shared among the participating clients in the federation, while keeping the data decentralized. In medical research, by bringing different hospitals and medical centers into the federation, researchers can collaboratively train a model utilizing different datasets from siloed institutions besides their own~\cite{andreux2020siloed,roth2020federated,wang2020automated,sarhan2020fairness,qayyum2021collaborative}.

However, the federated settings generate a new major challenge, namely the statistical data heterogeneity across different participating clients ~\cite{li2020federated,sahu2018convergence,li2018federated,karimireddy2020scaffold,ghosh2020efficient}. The data heterogeneity reflects that the data collected by different clients is not identically distributed (non-IID), which often appears in medical datasets from different sites, because of various reasons including different data acquisition protocols and different local demographics~\cite{rieke2020future}. Data heterogeneity may lead to significant increase in communication rounds of the federated training, and inferior performance of the distributed optimization of federated models in certain clients (e.g. medical institutes)~\cite{sahu2018convergence}, which can further cost their incentives to participate in the federation.

In this work, we propose a federated learning algorithm for classification tasks, \textit{Federated Learning with Shared Label Distribution} (\ct{FedSLD}), which aims to utilize information regarding the clients' label distribution, to estimate a general prior label distribution for the entire federation. We claim that \ct{FedSLD} can mitigate instability of training caused by the statistical heterogeneity of cross-silo FL, such as for medical research. While the algorithm does not access the clients' data, we assume legitimate for the clients to share the number of samples in each class, which are often the case for cross-silo FL such as in medical applications. More specifically, our contribution in this work is two-fold:

\begin{enumerate}[label=\textit{\roman*})]
    \item We propose a new FL algorithm for medical image classification: Federated Learning with Shared Label Distribution (\ct{FedSLD}), for robust training with non-IID data.
    \item We demonstrate that the proposed \ct{FedSLD} achieves better performance than the leading FL algorithms by conducting extensive experiments on four publicly available datasets (including two benchmark datasets) under pathological non-IID and practical non-IID data partitions.
\end{enumerate}

The rest of the paper is organized as follows: Section \ref{Method} provides details of the proposed \ct{FedSLD}; experiments and results are shown in Section \ref{Experiments and Results}; Section \ref{Conclusion} discusses our findings and concludes the study.

% \section{Related Work}
% \label{Related Work}
% Federated learning XXX (Fig about \ct{FedAvg}, central server)

% Federated learning with non-IID data XXX

% In medical research, FL XXX.

\section{Method}
\label{Method}
Laws and restrictions in terms of the data privacy constrain the direct access to the raw data. Yet, there are other information regarding the dataset that can be shared in terms of the federated learning. For instance, \ct{FedAvg} assumes knowledge of the number of samples in each client: after the aggregation step in \ct{FedAvg}, the algorithm conducts a weighted average of the updated copies for the next round, and the weights used for the averaging, by default, are the normalized number of samples in each client.

In this work, we focus on the classification tasks and assume legitimate to gain knowledge of the label distribution of each client, namely the number of samples from every class. We compute an estimate of the prior label distribution for the entire federation using the gain knowledge on the label distributions. For FL in medical applications, the label distributions from different medical silos can often be drastically different due to the regional demographics. Knowledge of the clients' label distributions will help us better understand the non-IID data in the federation.

To formulate the process, let us consider a federation with non-IID data. For a given data sample $(x,y)$, where $x$ stand for the data and $y$ represents the label, the probability that it appears in the dataset of client $i$'s, $\mP_i(x, y)$, does not necessarily equal to the probability of it to appear in the dataset of client $j$'s, $\mP_j(x,y)$. By Bayes' theorem, we have $\mP_i(x|y)\mP_i(y) \neq \mP_j(x|y)\mP_j(y)$. More often than not, especially in medical imaging domain, non-IID data implicitly implies that both the label-conditioned probabilities, $\mP(x|y)$, and the marginal label distributions, $\mP(y)$, are different for different clients. In this work, we focus on acquiring the information reflecting the marginal label distribution $\mP_i(y)$ for each client ($i=1,2,...,N$), to compute the estimate of the prior label distribution for the entire federation.

We define the estimate for the prior of class $c$ for the federation, as the sum of the numbers of samples for class $c$ in each client divided by the sum of the total number of samples in each client. This is shown in equation (\ref{prior}), where $\tP_(y=c)$ is the estimate prior of class $c$, $n_{i, c}$ is the number of samples from class $c$ on client $i$, $n_i$ is the total number of samples on client $i$, and $N$ is the number of clients.

\begin{equation}
    \tP(y=c) = \frac{\sum_{i=1}^N n_{i,c}}{\sum_{i=1}^N n_i}
    \label{prior}
\end{equation}

During local update of the current model on a client, given a batch of data $\{ (x_k, y_k) \}_{k=1}^B$, where $B$ is the batch size, we first compute the label distribution in this batch as in equation (\ref{batch_dist}), where the $p_b$ represents the label distribution, $\llbracket \cdot \rrbracket$ means the indicator faction, with its value equal to 1 if the inner part is true, and 0 otherwise. In essence, Equation (\ref{batch_dist}) computes the proportion of class $c$ samples in the batch by normalizing the number of class $c$ samples in this batch.

\begin{equation}
    p_b(y=c) = \frac{\sum_{k=1}^B \llbracket y_k=c \rrbracket}{B}
    \label{batch_dist}
\end{equation}

\begin{equation}
    \mathcal{L}_b\left(\big\{ (x_k, y_k) \big\}_{k=1}^B \right) = -\sum_{k=1}^{B} \left(\frac{p_b(y=y_k)}{\tP(y=y_k)} \cdot \sum_{c=1}^C y_{k,c} \log \left(f_i(x_k)\right)_c \right)
    \label{loss}
\end{equation}

Then, we define the batch loss as a weighted cross-entropy loss, shown in Equation (\ref{loss}), where $\mathcal{L}_b$ means the batch loss, and $f_i$ represents the copy of the model on client $i$. By doing this, we can enforce proportional contribution (to the local objective) of each class of the data, with respect to its share of the true underlying distribution across the federation. 

\begin{algorithm}[t]
    \caption{\textbf{FedSLD}.}\label{alg}
    \textbf{Input}: Initialized model parameter weights $w^0$, number of clients $N$, number of local epochs $E$, batch size $B$, is the batch size, learning rate $\eta$, number of rounds $R$.
    \begin{algorithmic}[1]
        \State $\forall i \in [N], c \in [C]$, acquire $n_{i,c}$, client $i$'s numbers of samples of each class $c$.
        \State $\forall c \in [C]$, $\tP(y=c) = \frac{\sum_{i=1}^N n_{i,c}}{\sum_{i=1}^N n_i}$ // compute estimated prior label distribution.
        \For{$r \leftarrow 1,2,...,R$}
            \State $\forall i \in [N]$ $w_i^r=w^{r-1}$ // broadcast the model parameters.
            \For{$i \leftarrow 1,2,...,N$ \textbf{in parallel}}
                \For{$\{x_k, y_k\}_{k=1}^B$ \textbf{in all minibatches}}
                    \State $\forall c, \ p_b(y=c) \gets \sum_{k=1}^B \llbracket y_k=c \rrbracket / B$
                    \State $\mathcal{L}_b \gets -\sum_{k=1}^{B} \left(\frac{p_b(y=y_k)}{\tP(y=y_k)} \cdot \sum_{c=1}^C y_{k,c} \log \left(f_i(x_k)\right)_c \right)$
                    \State $w_i^r \gets w_i^{r} - \eta \nabla_w\mathcal{L}_b$
                \EndFor
            \EndFor
            \State $w^r = \sum_{i=1}^N \frac{n_i}{n} w_I^r$ // aggregate the model updates
        \EndFor
        \State \textbf{return} $w^R$
    \end{algorithmic}
\end{algorithm}

We follow the aggregation step in a typical FL algorithm such as \ct{FedAvg}, where we compute the weighted average of the updated models from all clients, with the weights being the number of samples in each client. A detailed algorithm is shown in Algorithm \ref{alg}.

\section{Experiments and Results}
\label{Experiments and Results}
In this section, we evaluate the performance of the proposed \ct{FedSLD} through experiments on four publicly available datasets (including two benchmark imaging datasets), and compare it with two leading FL algorithms, \ct{FedAvg}~\cite{mcmahan2017communication}, an algorithm that average the local updates of the global model, and \ct{FedProx}~\cite{li2018federated}, an algorithm that adds a proximal term on the local objective to enhance performance robustness on non-IID data. To evaluate the general performance of the algorithms, we compute the test accuracies and demonstrate the empirical convergence performance by plotting the training loss and test accuracy curves. In addition, we examine the fairness of the method following recent work~\cite{li2019fair}. More details on the metrics are in Section \ref{Metrics}.

\subsection{Experiments setup}
\textbf{Datasets. }
\label{datasets}
We conduct experiments on two benchmark image datasets: MNIST ~\cite{lecun-mnisthandwrittendigit-2010}, a 10-class handwritten digit classification dataset and CIFAR10~\cite{krizhevsky2009learning}, a 10-class dataset with animals and transportations images. We further evaluate the methods on two medical image datasets from the MedMNIST dataset collection~\cite{yang2021medmnist}, namely the OrganMNIST(axial) dataset: an 11-class dataset of liver tumor images~\cite{bilic2019liver}, and the PathMNIST dataset: a 9-class dataset of colorectal cancer images~\cite{kather2019predicting}. We partition each dataset into a training set and a test set and ensure that they share the same label distribution.

\begin{figure}[t]
    \centering
    \includegraphics[width=1\textwidth]{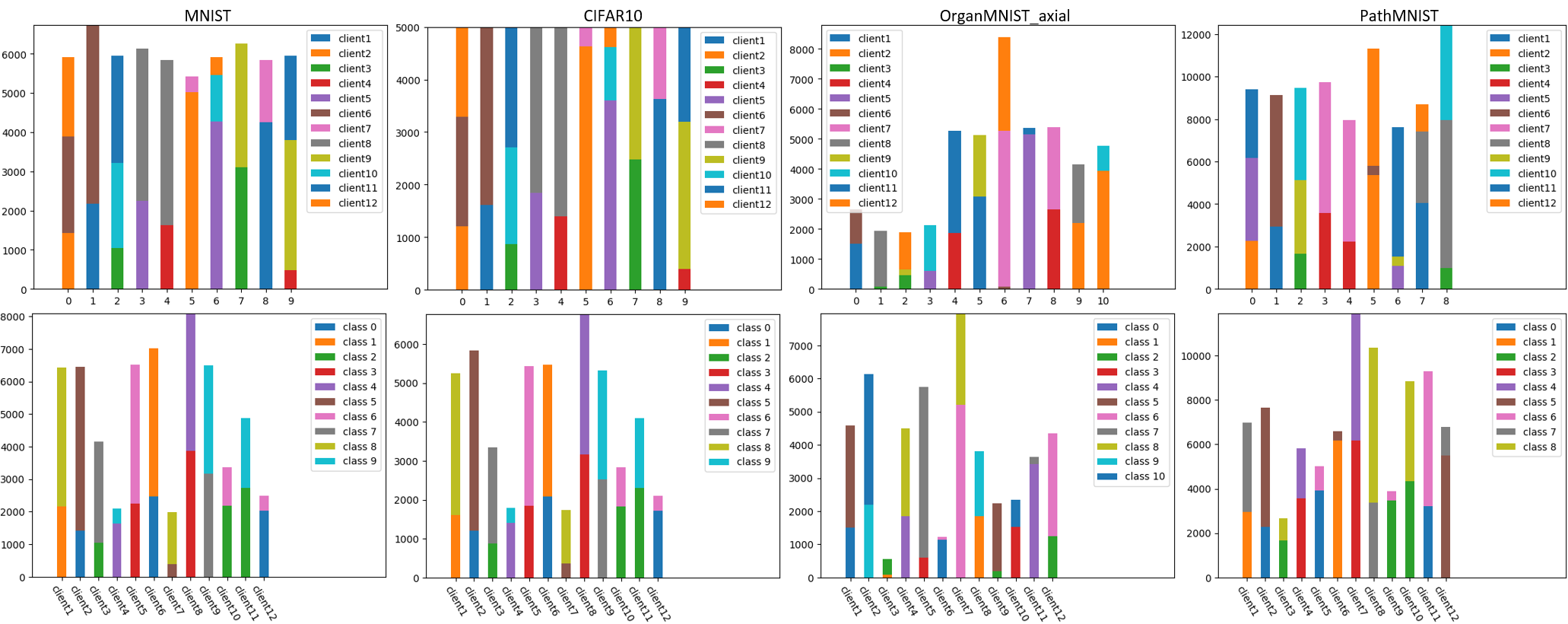}
    \caption{Pathological non-IID distribution. The first row depicts the portions locations from each class. The second row depicts the dataset composition of each client.} \label{path_data_dist}
\end{figure}

\begin{figure}[ht]
    \centering
    \includegraphics[width=1\textwidth]{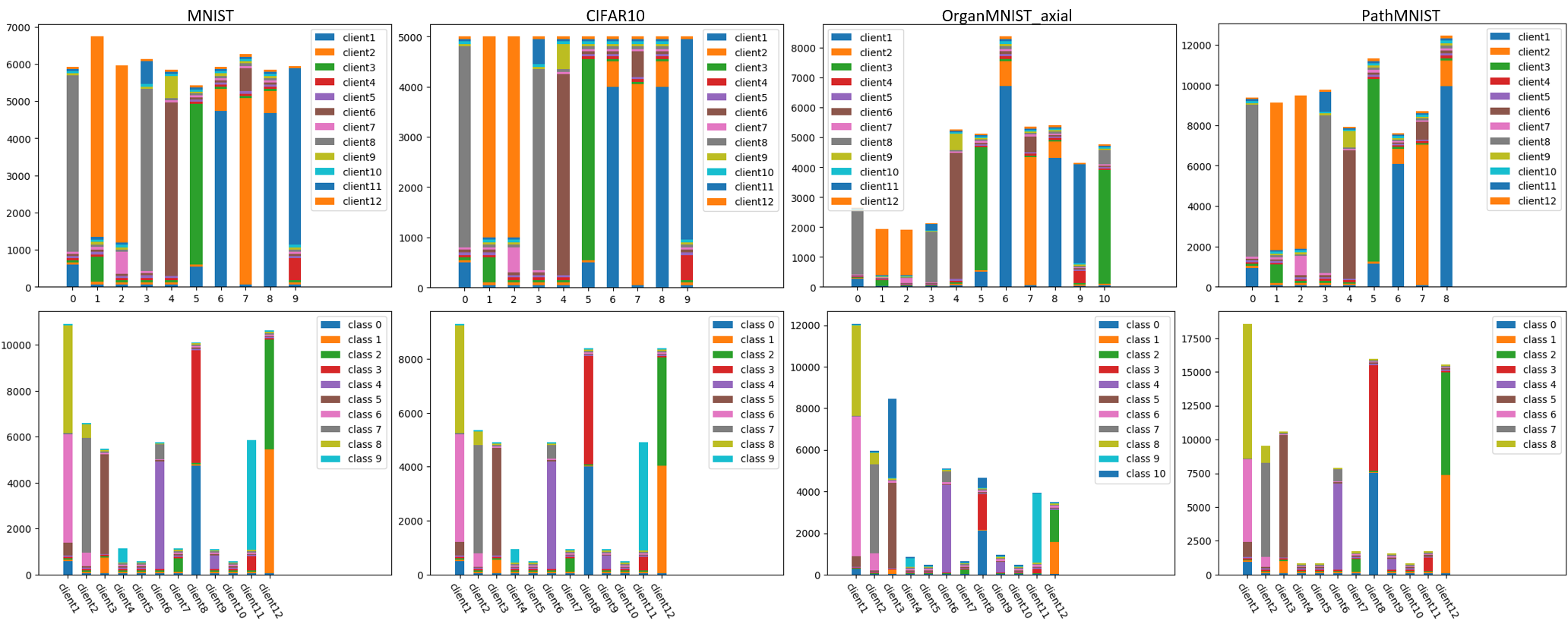}
    \caption{Practical non-IID distribution. The first row depicts the portions' locations from each class. The second row depicts the dataset composition of each client.} \label{prac_data_dist}
\end{figure}

\textbf{Two non-IID settings. }
\label{non-IID}
We partition each dataset according to two different non-IID settings: 1) a pathological non-IID setting. In this setting, we follow~\cite{mcmahan2017communication} by assigning each client with two random classes. A random number of images from these two classes are assigned to this client. We set the number of clients to be 12 to mimic a cross-silo FL setting. Figure \ref{path_data_dist} shows the pathological non-IID setting in more details; 2) a practical non-IID setting. In this setting, we randomly partition each class in the training set into 12 shards (corresponding to a total of 12 clients in the federation): $10$ shards of $1\%$, one shard of $10\%$ and one shard of $80\%$. For each client, we randomly assign a shard from each class to this client, so that the client will possess images from all classes, with more images from some classes while less images from others. This non-IID setting is more similar to the real-world medical applications, since datasets held at medical centers often contain a variety of classes, but medical centers in different regions, due to local demographics, may present different occurrence of different classes. Consequently, the datasets at the medical centers are often imbalanced with different majority classes. Figure \ref{prac_data_dist} shows the data distribution of all four datasets in more details.

% For both non-IID settings, we partition the test set with the same exact partition (in percentage) as the training set for the clients, so that for each client, its training set and test set share the same distribution.

\textbf{Implementation details.}
We use the classic four-layer CNN model with two 5x5 convolutional layers and two fully connected layers (hidden layer has 500 units). We use a batch size of 256, 5 local epochs, 0.01 as the learning rate. For the practical non-IID partition, we train the model for 80 rounds, and for the pathological non-IID setting, we train the model for 160 rounds. All experiments are run on an NVIDIA Tesla V100 GPU and implemented in PyTorch~\cite{NEURIPS2019_9015} and PySyft~\cite{ryffel2018generic}.

\textbf{Metrics.}
\label{Metrics}
We compute two types of test accuracies for each setting. 1) The \textit{Best Mean Client Test Accuracy} (BMCTA), referred as BMTA in~\cite{huang2021personalized}. BMCTA is computed as the highest mean client test accuracy of each round. 2) The \textit{Best Test Accuracy} (BTA). We treat the test sets from different clients as a combined test set, and compute the highest test accuracy over all round. We also investigate the methods' convergence performance by plotting the training loss and test accuracy curves. In addition, we follow~\cite{li2019fair} and examine the fairness of the methods by using the Gaussian kernel density estimation on the client test accuracies. Higher density at higher accuracy reflects a better result.

\begin{table}[t]
    \caption{The \textit{Best Mean Client Test Accuracy} (BMCTA) and \textit{Best Test Accuracy} (BTA) for the pathological non-IID setting}\label{path_TA}
    \centering
    \begin{tabular}{wl{3cm}M{1.4cm}M{1.4cm}M{1.4cm}M{1.4cm}M{1.4cm}M{1.4cm}}
    \toprule
    & \multicolumn{3}{c}{BMCTA}         &    \multicolumn{3}{|c}{BTA}    \\
    \multirow{2}{*}{Dataset}  &  \ct{FedAvg} & \ct{FedProx} & \ct{FedSLD} & \multicolumn{1}{|c}{\ct{FedAvg}} & \ct{FedProx} & \ct{FedSLD}\\ 
    & ~\cite{mcmahan2017communication} & ~\cite{li2018federated} & (Ours) & \multicolumn{1}{|c}{~\cite{mcmahan2017communication}} & ~\cite{li2018federated} & (Ours)\\\midrule
    MNIST & 95.60 & 95.71 & \textbf{95.74} & \multicolumn{1}{|c}{95.92} & 95.98 & \textbf{96.03} \\
    CIFAR10 & \textbf{51.50} & 51.39 & 50.81 & \multicolumn{1}{|c}{\textbf{51.39}} & 51.24 & 50.71 \\
    OrganMNIST(axial) & 59.52 & 59.44 & \textbf{59.70} & \multicolumn{1}{|c}{64.99} & 65.10 & \textbf{66.13} \\
    PathMNIST & 56.44 & 56.62 & \textbf{57.94} & \multicolumn{1}{|c}{57.54} & 57.56 & \textbf{59.11} \\ \bottomrule
    \end{tabular}
\end{table}

\begin{figure}[ht]
    \centering
    \includegraphics[width=1\textwidth]{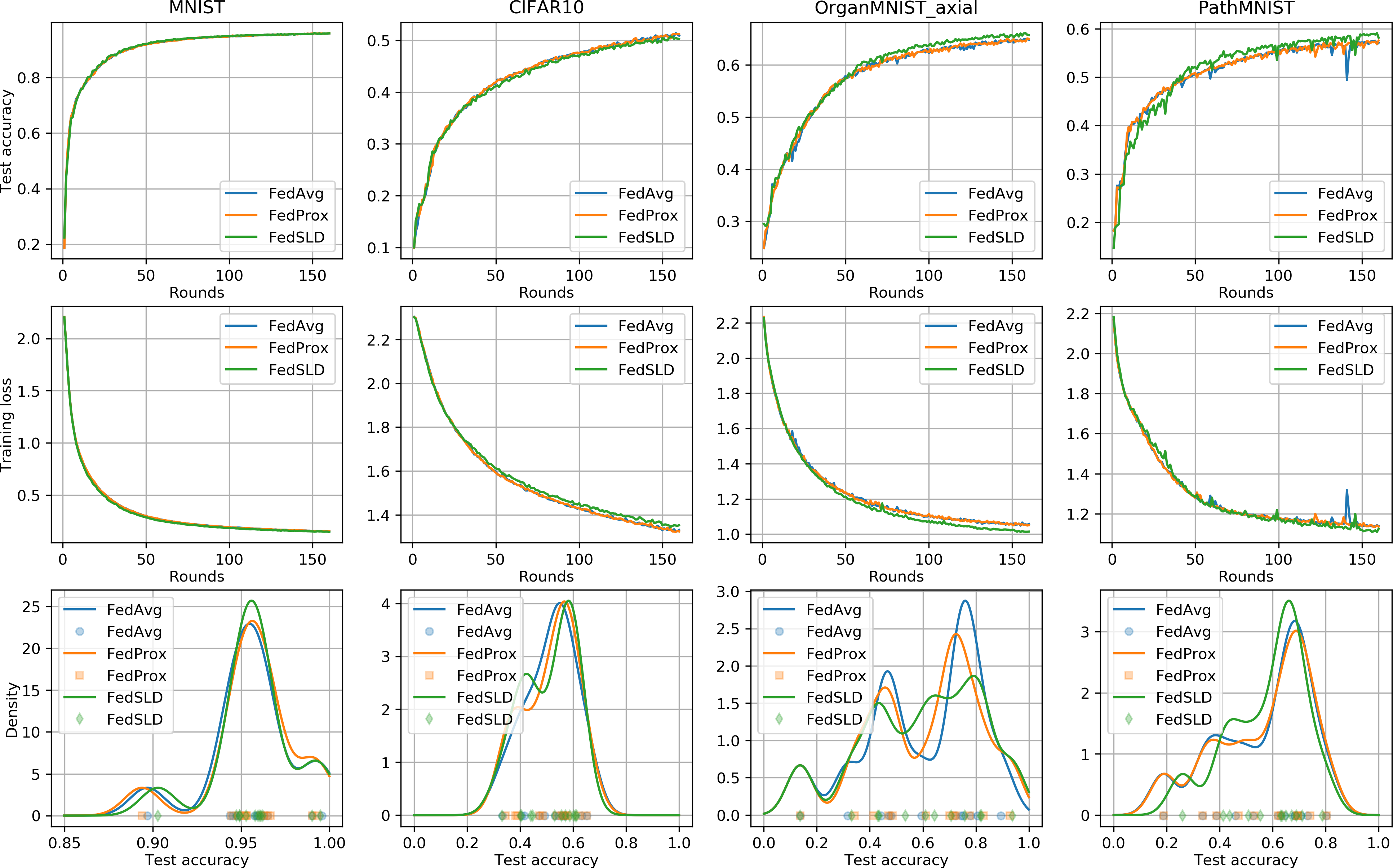}
    \caption{The convergence and fairness performance under the pathological non-IID setting. We measure the fairness using Gaussian kernel density estimation. Higher density concentrated at a higher accuracy reflects a better result.} \label{path_plots}
\end{figure}

\subsection{Results}
Under the pathological non-IID setting, Table \ref{path_TA} and Fig. \ref{path_plots} show that for MNIST, and the two medical datasets, the proposed \ct{FedSLD} has a better performance with the improvement on the test accuracy of up to 1.57\%, and the kernel density estimations show that \ct{FedSLD} has slightly higher density which is more concentrated at a higher test accuracy. On CIFAR10, \ct{FedSLD} reaches competitive performance with \ct{FedAvg} and \ct{FedProx}.

\begin{table}[h!]
    \caption{BMCTA and BTA for the practical non-IID setting}\label{prac_TA}
    \centering
    \begin{tabular}{wl{3cm}M{1.4cm}M{1.4cm}M{1.4cm}M{1.4cm}M{1.4cm}M{1.4cm}}
    \toprule
    & \multicolumn{3}{c}{BMCTA}         &    \multicolumn{3}{|c}{BTA}    \\
    \multirow{2}{*}{Dataset}  &  \ct{FedAvg} & \ct{FedProx} & \ct{FedSLD} & \multicolumn{1}{|c}{\ct{FedAvg}} & \ct{FedProx} & \ct{FedSLD}\\ 
    & ~\cite{mcmahan2017communication} & ~\cite{li2018federated} & (Ours) & \multicolumn{1}{|c}{~\cite{mcmahan2017communication}} & ~\cite{li2018federated} & (Ours)\\\midrule
    MNIST & 93.41 & 93.45 & \textbf{95.56} & \multicolumn{1}{|c}{94.15} & 94.20 & \textbf{95.85} \\
    CIFAR10 & 32.07 & 31.98 & \textbf{37.48} & \multicolumn{1}{|c}{35.46} & 35.38 & \textbf{37.79} \\
    OrganMNIST(axial) & 82.32 & 81.53 & \textbf{84.75} & \multicolumn{1}{|c}{85.69} & 85.54 & \textbf{87.37} \\
    PathMNIST & 52.70 & 52.77 & \textbf{53.87} & \multicolumn{1}{|c}{57.38} & 57.72 & \textbf{57.90} \\ \bottomrule
    \end{tabular}
\end{table}

\begin{figure}[t]
    \centering
    \includegraphics[width=1\textwidth]{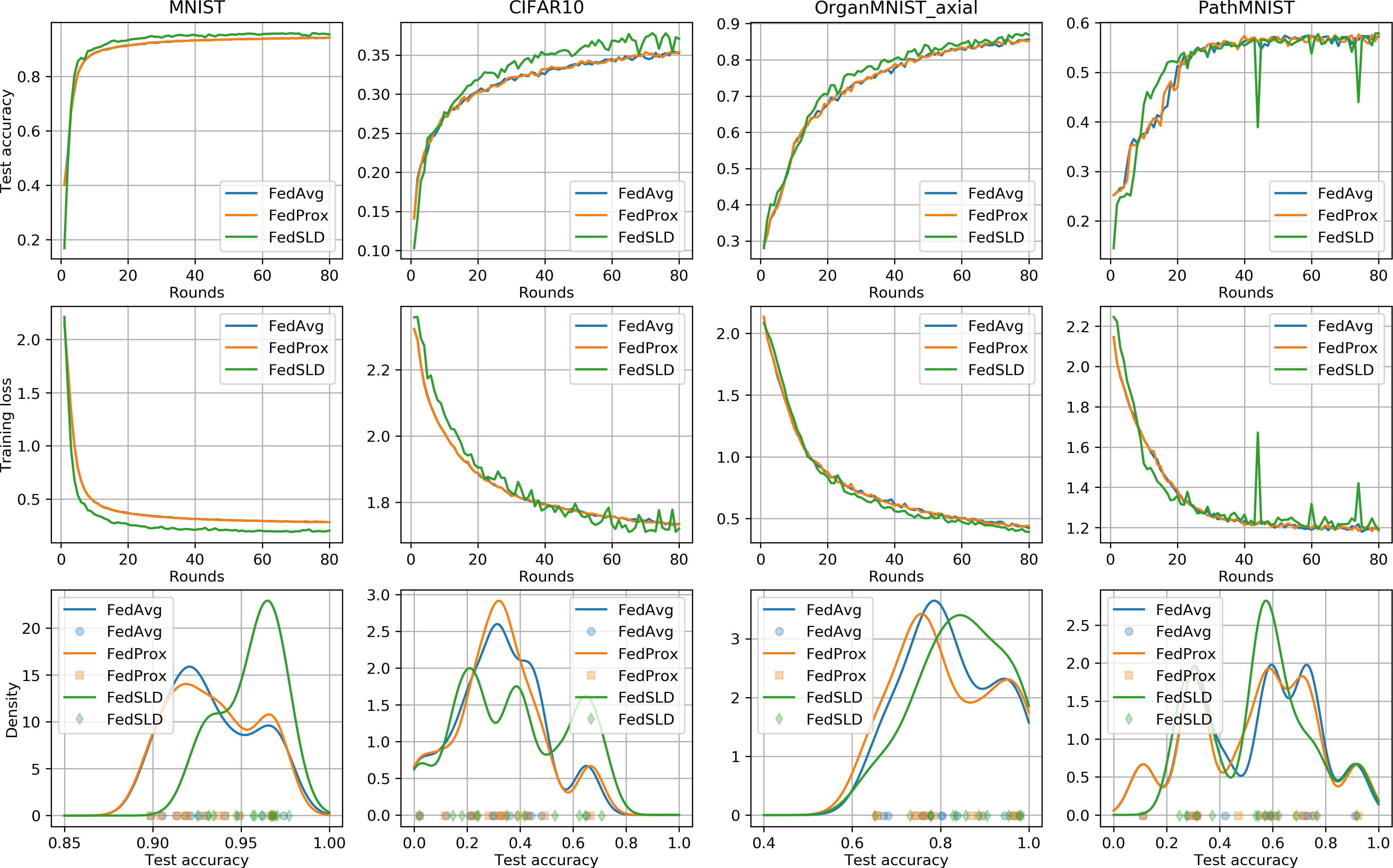}
    \caption{The convergence and fairness performance under the practical non-IID setting.} \label{prac_plots}
\end{figure}

Under the practical non-IID setting, we can see from Table \ref{prac_TA} and Figure \ref{prac_plots} that the proposed \ct{FedSLD} outperforms the compared \ct{FedAvg} and \ct{FedProx} on every dataset, with the improvement of BMCTA ranging from 1.10\% to 5.50\%, and the improvement of BTA ranging from 0.18\% to 2.41\%. In addition, \ct{FedSLD} achieves better convergence behavior on MNIST and OrganMNIST (axial) datasets. The fairness plots reveal that \ct{FedSLD} not only increases the overall performance with respect to the entire federation, but the variances of the client test accuracies are also reduced on MNIST and PathMNIST datasets, which implies a more fair training. On CIFAR10 and OrganMNIST (axial) datasets, we can see a clear decrease of the density at low accuracy and an increase on the density at high accuracy, which explains the improvement of the BMCTA.

\section{Conclusion}
\label{Conclusion}
In this work, we proposed a new FL algorithm for medical image classification: Federated Learning with Shared Label Distribution (\ct{FedSLD}). \ct{FedSLD} aims to mitigate the effect caused by non-IID data by leveraging the clients' label distribution. We conducted extensive experiments on four publicly available datasets with two types of non-IID setting, and demonstrated that \ct{FedSLD} outperforms the compared leading FL algorithms, and encourages a more fair performance across all the participating clients.

\section{Compliance with ethical standards}
\label{sec:ethics}
Ethical approval was not required, as this study used previously collected and deidentified data (including medical imaging data) available in public repositories.

\paragraph{\textbf{Acknowledgements.}}
This work was supported in part by National Institutes of Health (NIH) (1R01CA193603 and 1R01CA218405), National Science Foundation (NSF) (CICI: SIVD: 2115082), and the grant 1R01EB032896 as part of the NSF/NIH Smart Health and Biomedical Research in the Era of Artificial Intelligence and Advanced Data Science Program. This work used the Extreme Science and Engineering Discovery Environment (XSEDE), which is supported by NSF grant number ACI-1548562. Specifically, it used the Bridges-2 system, which is supported by NSF award number ACI-1928147, at the Pittsburgh Supercomputing Center.

\bibliographystyle{splncs04}
\bibliography{mybibliography.bib}

\end{document}